\documentclass[conference]{IEEEtran}
 
\newcommand\subparagraph{%
  \@startsection{subparagraph}{5}
  {\parindent}
  {3.25ex \@plus 1ex \@minus .2ex}
  {-1em}
  {\normalfont\normalsize\bfseries}}
\usepackage{cite}
\usepackage{amsmath,amssymb,amsfonts}
\usepackage{algorithmic}
\usepackage{graphicx}
\graphicspath{{./images/}}
\usepackage{textcomp}
\usepackage{xcolor}
\usepackage[font=footnotesize]{caption}
\usepackage{subcaption}
\usepackage{multirow}
\usepackage{rotating}
\usepackage{subcaption}
\usepackage{todonotes}
\usepackage{titlesec}

\titlespacing{\section}{0pt}{*0.5}{*0.5}
\titlespacing{\subsection}{4pt}{*0.5}{*0.5}
\begin{document}
\title{\LARGE \bf Gesture Recognition for Initiating Human-to-Robot Handovers}
\author{\IEEEauthorblockN{Jun Kwan, Chinkye Tan and Akansel Cosgun}
    
\IEEEauthorblockA{Monash University, Australia}
}

\maketitle

\begin{abstract}
Human-to-Robot handovers are useful for many Human-Robot Interaction scenarios. It is important to recognize when a human initiates handovers, so that the robot does not try to take objects from humans when a handover is not intended. We pose the handover gesture recognition as a binary classification problem in a single RGB image. Three separate neural network modules for detecting the object, human body key points and head orientation, are implemented to extract relevant features from the RGB images, and then the feature vectors are passed into a deep neural net to perform binary classification. Our results show that the handover gestures are correctly identified with an accuracy of over 90\%. The abstraction of the features makes our approach modular and generalizable to different objects and human body types.

\end{abstract}

\section{Introduction}

Robots are under rapid development in sectors such as manufacturing, automation and hospitality. Collaborative robots are increasingly being used in industry, and is expected to be useful in home environments in the future. One of the expected capabilities for collaborative robots is the ability to perform \textit{object handovers}. Object handovers can happen in two directions: Robot-to-Human where the robot delivers a requested object to a human, and Human-to-Robot where the robot acquires an object from a human. The object handover problem has been studied extensively in the robotics literature, with most of the published work focusing on the Robot-to-Human handover scenario. Human-to-Robot handover scenario is arguably the more difficult problem because, as opposed to the Robot-to-Human scenario, the robot responsible for grasping the object from the human partner's hand, which raises safety issues. This makes the perception problem very critical for the Human-to-Robot handover scenario. The perception system is responsible for detecting when and where a human partner wants to engage in a handover, detect the object the human wants to pass, and ensure that a finger or body part of the human is not grasped. For example, a human might be holding a cellphone near the robot, and the robot should not unexpectedly reach out and grasp it from the human's hand. In this paper, we focus on detecting if a human is initiating a Human-to-Robot object handover.

\begin{figure}[ht!]
    \centering
    \begin{subfigure}{0.49\columnwidth} 
        \includegraphics[trim={0.1cm 0.1cm 5cm 0.1cm},clip,width=\textwidth]
            {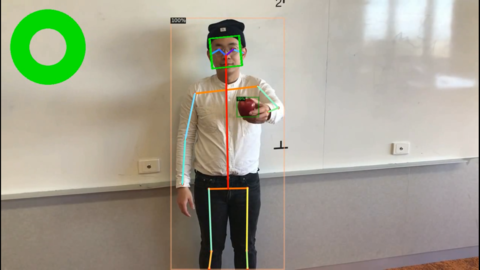}
        \label{fig:reg1}
    \end{subfigure}
    \begin{subfigure}{0.49\columnwidth} 
        \includegraphics[trim={0.1cm 0.1cm 5cm 0.1cm},clip,width=\textwidth]
            {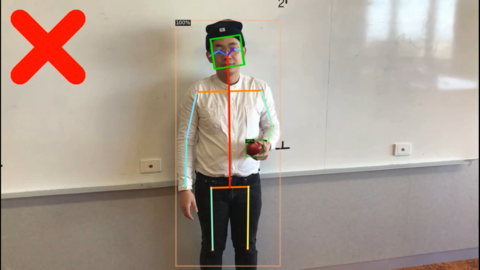}
        \label{fig:reg2}
    \end{subfigure}
    % \begin{subfigure}{0.49\columnwidth} 
    %     \vspace{-10pt}
    %     \includegraphics[trim={0.1cm 0.1cm 5cm 0.1cm},clip,width=\textwidth]
    %         {images/intro3.png}
    %     \label{fig:reg3}
    % \end{subfigure}
    % \begin{subfigure}{0.49\columnwidth}
    %     \vspace{-10pt}
    %     \includegraphics[trim={0.1cm 0.1cm 5cm 0.1cm},clip,width=\textwidth]
    %         {images/intro4.png}
    %     \label{fig:reg4}
    % \end{subfigure}
    \vspace{-10pt}
    \caption{We detect whether a user is initiating a handover from a single RGB image. Handover gesture is correctly detected (left column). No handover activity is detected  (right column).}
    \label{fig:holds}
\end{figure}

Human-to-Robot handovers have been demonstrated on a physical robot system by a handful of researchers \cite{Yamane2013, Maeda2016, Konstantinova2017, Pan2018, Yang2020, Rosenberger2020}. Even though successful demonstrations has been shown, in these works the robot is programmed for a single task: object handovers. In real scenarios where the robot is expected to engage in many tasks, the robot must first recognize the action of the human handing an object over to the robot. There are many communication cues that humans use to recognize handover intent, including direct cues such as verbal communication, and indirect cues such as eye gaze and body gestures. In this work, we focus on the binary classification of whether a human partner is handing over an object from body gestures only. 

Our approach is based on extracting features relevant to the task from three independent modules, and learning a classifier to detect the existence of a handover gesture, based on the extracted features. Each of these modules is a neural network, described below:
\begin{itemize}
\setlength\itemsep{0pt}
    \item Object Detector: Detect the presence and pixel coordinates of the bounding box of an object.
    \item Human Body Pose Detector: Detect the human body pose with specific key points.
    \item Head Orientation Detector: The head orientation of the person, represented in Euler angles with respect to the camera frame.
\end{itemize}

We train a deep neural network that converts the resulting feature vector generated by the modules into a binary classification result. Our system uses the Faster Region Based Convolutional Neural Network (Faster R-CNN) \cite{ren2015faster} for object detection, Keypoint R-CNN \cite{he2017mask} to estimate the human body pose, multi-loss Resnet50 \cite{ruiz2018fine} architecture to estimate the head pose of a person, and a fully connected deep net to produce the final result for recognition. We represent the human body key points in the object coordinate frame so that the detection is independent of where the human is in the image frame. Our approach is 1) \textit{modular} because each module can be replaced as long as the output type is the same 2) \textit{generalizable} because it is independent of the object and human appearance, and the relative position of the human in the image.

\section{Approach}

\begin{figure}[ht!]
\centering
\includegraphics[scale=0.5]{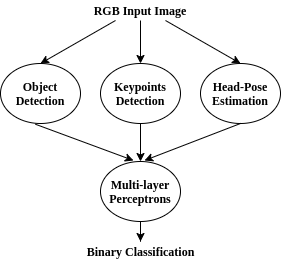}
\caption{System Diagram}
\vspace{-10pt}
\label{fig:system_diagram}
\end{figure}

Our approach utilizes a total of four modules as shown in the system diagram (Fig.~\ref{fig:system_diagram}). The input RGB image is passed into the three neural networks to obtain relevant features in parallel: An object detector detects the pixel coordinates of the object of interest, multiple keypoints of a person's body is detected, and the head pose of the person is estimated. The resulting features are processed and compressed into a feature vector, which is then passed into the final multi-layer perceptron (MLP) to generate a binary output, which is the estimation of whether the human is ready for a handover. 

\subsection{Object Detection}
Faster R-CNN\cite{ren2015faster} is implemented to detect pixel coordinates of an object held by a person. When an object is detected, a bounding box will be generated by this module. In the case where multiple objects are detected, the object with largest bounding box area is selected to be the target. Faster R-CNN is a modification of the original Region Based CNN\cite{girshick2014rich} and the improved Fast R-CNN\cite{girshick2015fast}. R-CNNs introduces a selective search algorithm to propose regions of interest in the image. These regions are then individually passed into a CNN which extracts the features of the regions. These features are subsequently passed into an SVM which determines the presence of the object in the proposed regions. Fast R-CNN improves the performance of R-CNN by passing the image through the CNN first to extract the feature map which is then used to generate the region proposals. Fast R-CNN is faster than basic R-CNN because the image is only passed into the CNN once whereas in R-CNN the large number of region proposals are all passed into the CNN slowing down the entire network. Faster R-CNN takes this improvement further by switching out the selective search algorithm used by R-CNN and Fast R-CNN to generate the region proposals with a region proposal network.

\subsection{Body Keypoints Detection}

Human body keypoints is commonly used in robotics applications, such as for ensuring safety for Human-Robot Interaction \cite{cosgun2013accuracy,Rosenberger2020} and recognizing gestures such as a pointing gestures \cite{cosgun2015did}.

Mask R-CNN\cite{he2017mask} is implemented to detect various keypoints of a human body, such as shoulders, elbows, wrists, etc. Mask R-CNN is a modification of Faster R-CNN, where in parallel to the class and bounding box prediction it also adds a branch that outputs a binary mask for each Region of Interest (RoI). Mask R-CNN is able to perform semantic segmentation on each separate RoI, allowing the model to identify the boundaries of objects at the pixel level. 

For keypoints detection, each keypoint is treated as a separate class during the training process. The segmentation branch of Mask R-CNN outputs \textit{k} binary masks representing \textit{k} different keypoints. In each binary mask, only one pixel is labelled as the foreground by the model representing the location of the keypoint. The coordinates of each of these points are then taken as the estimation of the pose. This modification of the Mask R-CNN model is called Keypoint R-CNN. 

\subsection{Head Pose Estimation}
Multi-loss Resnet50 architecture, proposed by Ruiz et al.\cite{ruiz2018fine}, is utilised for head pose estimation. Multiple losses are involved in the work and a loss is designated for each Euler angle. Thus, there are a total of three losses which are designated for yaw, pitch and roll separately. A binned pose classification and a regression loss are included in each loss. Furthermore, ResNet50 is used as the backbone of the network and three fully connected layers are attached to it for Euler angle predictions. A softmax layer and a cross entropy loss are implemented for bin classification, and hence there are three cross entropy losses in total. These losses are then back-propagated through the network. A mean squared error loss is added to the network as a regression loss as well. In addition, multi-task cascaded convolution networks (MTCNN)\cite{zhang2016joint}, proposed by Zhang et al., is used for face detection before computing the head pose estimation of the person.

\subsection{Gesture Recognition from Extracted Features}
\label{sec:gesture_recognition}
\begin{figure}[ht!]
\centering
\includegraphics[scale=0.5]{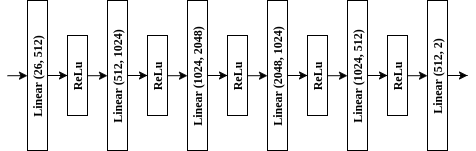}
\caption{Neural network architecture for the MLP to detect handover gestures}
\label{fig:mlp}
\end{figure}

We used a Multi-Layer Perceptron that consists of an input layer, four hidden layers, and an output layer. The network architecture is illustrated in Fig.~\ref{fig:mlp}. Features extracted by the previous modules are processed and concatenated into a feature vectore before passing into the MLP. Only keypoints of upper body of a person are included in the feature vector. We implemented two ways to input the feature vector, according to how the body keypoints are represented: relative to the object, or absolute pixel coordinates. Below we explain each method.

\begin{itemize}
\setlength\itemsep{0pt}
    \item \textbf{Absolute Pixels: } The input feature vector has a size of 29. The following features are concatenated into a vector.
    \begin{itemize}
        \item Object (4 values): The pixel coordinates of the centroid, width and height of the object bounding box.
        \item Absolute Human body keypoints (22 values): Pixel coordinates of each of the 11 keypoints belonging to upper body.
        \item Head orientation (3 values): Yaw, pitch and roll angles in degrees, represented in camera frame.
    \end{itemize}
    \item \textbf{Relative to Object: } The input feature vector has a size of 26. The following features are concatenated into a vector.
        \begin{itemize}
        \item Object presence (1 value): Binary value of whether an object of interest is detected.
        \item Relative human body keypoints (22 values): Pixel coordinates of each of the 11 keypoints belonging to upper body, represented in the 2D coordinate frame attached to the target object centroid.
        \item Head orientation (3 values): Yaw, pitch and roll angles in degrees, represented in camera frame.
    \end{itemize}
\end{itemize}

The output of the MLP is passed through a sigmoid function resulting in a likelihood of gesture detection between [0,1], and compared against a threshold (set to 0.5 in current implementation) to obtain the result of binary classification. 

\section{Experiments}

\subsection{Datasets and Training}

For object detection model, the Faster R-CNN is implemented with Facebook AI Research's Detectron2 engine \cite{wu2019detectron2}. Detectron2 engine contains a collection of state-of-the-art object detection algorithms as well as APIs to train and deploy multiple algorithms at ease. The Faster R-CNN model is loaded with the COCO 2017 pre-trained weights and retrained on a subset of the COCO 2017 dataset \cite{lin2015coco}. For this implementation, the object considered in the handover process is an apple, and the Faster R-CNN model is retrained on just the images of apples in the COCO 2017 dataset. Although the system is trained to recognize apples only, this choice is made arbitrarily given the objects at hand. Even though our system is implemented for a single object class, it will also work on generic object detectors as long as the object bounding box is generated.

For the body keypoints detection, the Keypoint R-CNN model is also implemented using the Detectron2 engine \cite{wu2019detectron2}. The model uses the pre-trained weights provided by the engine itself. The pre-trained weights are found to be sufficiently accurate at detecting the human poses usually encountered in a handover scenario.

For head pose estimation, we used the 300W across Large Poses (300W-LP) dataset \cite{zhu2016face} to train the network. The pre-trained model is loaded during the training phase. This dataset is initially used for 3D Dense Face Alignment (3DDFA)\cite{zhu2016face}, whereby a convolutional neural network (CNN) is used to apply a dense 3D model to an image. Moreover, 300W-LP contains various 3D landmarks which will be useful for training to obtain Euler angles such as yaw, pitch and roll.

For the handover gesture detection, we created a custom dataset for training the MLP model. A total of 25 videos were recorded in a lab setting, containing a total of 2506 images. Each image was given a label denoting a handover scenario or otherwise. For handover scenarios, the image was given a label 1 and 0 for non-handover scenarios. The images were individually labelled by hand. We use a rough 80\%/20\% training and testing split for training, in which while enforcing balancing between positive and negative examples. We also store all features extracted from RGB images in JSON format.

The training process begins with slicing a raw RGB video input into images and feeding them into the object detection, body pose estimation, and head pose estimation modules sequentially. The respective feature outputs are then processed for the MLP model. At this stage, the feature outputs are vectorised and passed into the MLP model. Some computation is also performed in this stage to make the frame to be relative to the object. There are a total of 26 parameters used for training essentially. When no object is detected, the yaw, pitch and roll in the feature vector are set to a dummy variable (-999 in this case) and the rest is set to be 0. This is to indicate an invalid feature. Binary Cross Entropy (BCE) loss is implemented in the training of MLP.

\subsection{Baselines}

In addition to our approach with two variations (absolute and relative human body key points), as explained in Sec.~\ref{sec:gesture_recognition}, we implemented two baselines:
\begin{itemize}
    \item \textbf{End-to-end:} Alexnet \cite{alex2012imagenet} and Resnet50 \cite{he2015deep} are used for this end-to-end image classification. A raw RGB image is provided as the input to the model to do feature extraction. Alexnet eventually classifies all images as 0 and hence it completely fails at the classification of the custom dataset. On the other hand, the accuracy of Resnet50 is significantly higher than Alexnet.
    \item \textbf{CNN on skeleton image:} Resnet50 \cite{he2015deep} is used for skeleton image classification. A skeleton image is generated by colouring the corresponding coordinates of each estimation point, bounding box, and head orientation on a black background. The image is then passed into Resnet50 to perform feature extraction. Resnet50 can only successfully classify some skeleton images in the custom dataset to a small extent.
\end{itemize}

\subsection{Quantitative Results}

The detection accuracy for our method (averaged over 5 random train/test splits), as well the baselines are shown in Table~\ref{table:metrics}.

\begin{table}[ht!]
\centering
\begin{tabular}{|c|c|}
\hline
\textbf{Method} &  \textbf{Accuracy (\%)} \\
\hline
End-to-end (Alexnet) & 50.0 \\
\hline
End-to-end (Resnet50) & 89.4 \\
\hline
CNN on skeleton image & 83.3 \\
\hline
Ours (absolute pixels) & 90.1 \\
\hline
\textbf{Ours (relative to object)} & \textbf{90.6} \\
\hline
\end{tabular}
\caption{Handover gesture detection accuracy. Our method, with human pose represented relative to object, performs the best.}
\label{table:metrics}
\end{table}

\begin{figure*}[ht!]
\centering
    \begin{subfigure}{\columnwidth} 
        \includegraphics[trim={0.1cm 0.1cm 4.5cm 0.1cm},clip,width=0.475\textwidth]
            {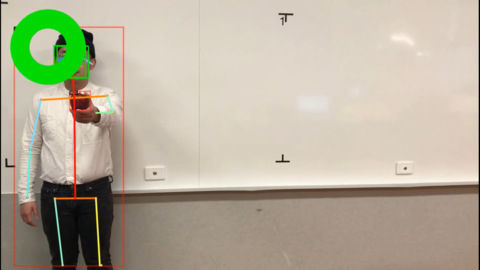}
        \includegraphics[trim={0.1cm 0.1cm 4.5cm 0.1cm},clip,width=0.475\textwidth]
            {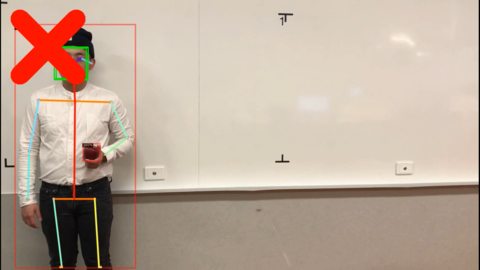}
        \caption{Standing at the corner}
        \label{fig:corner}
    \end{subfigure}
    \begin{subfigure}{\columnwidth} 
        \includegraphics[trim={0.1cm 0.1cm 4.5cm 0.1cm},clip,width=0.475\textwidth]
        {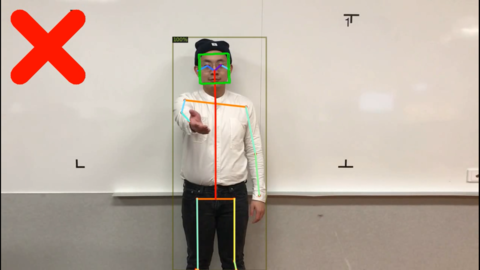}
        \includegraphics[trim={0.1cm 0.1cm 4.5cm 0.1cm},clip,width=0.475\textwidth]
        {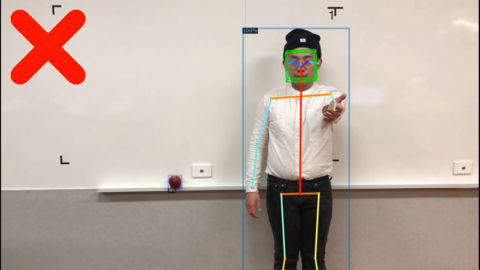}
        \caption{No object is detected (Left). Object is not in hand (Right).}
        \label{fig:unique_cases}
    \end{subfigure}
\vspace{-5pt}
\caption{Some successful examples of the classification results. X and O shows indicates the correct label for the image.}
\vspace{-15pt}
\label{fig:qualitative_results}
\end{figure*}

\begin{figure}[ht!]
\centering
    \begin{subfigure}{\columnwidth} 
        \includegraphics[trim={0.1cm 0.1cm 4.5cm 0.1cm},clip,width=0.475\textwidth]
        {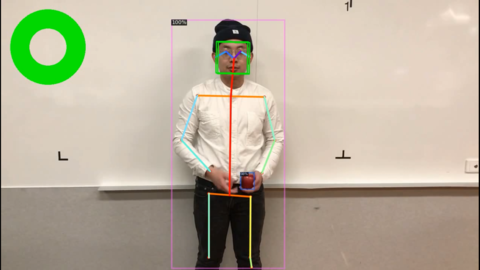}
        \includegraphics[trim={0.1cm 0.1cm 4.5cm 0.1cm},clip,width=0.475\textwidth]
        {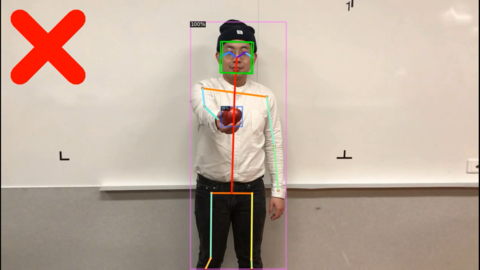}
    \end{subfigure}
\caption{Some failure cases. False positive (Left). False negative (Right)}
\label{fig:failure_cases}
\end{figure}

Our results show that the \textbf{End-to-End} method with Alexnet, which uses the raw RGB image, was no better than random guesses. This shows the difficulty of the problem, as the difference between positive and negative examples are very subtle. The \textbf{End-to-End} with Resnet50 and \textbf{CNN on Skeleton Image} approaches achieved 89.4\% and 83.3\% accuracy respectively. 

Both variations of our approach outperformed the baselines. Our method that uses the feature vector relative to the object had 90.6\% detection accuracy, performing slightly better than the feature representation that uses absolute pixel values.

\subsection{Qualitative Results}

Some qualitative results are shown in Fig.~\ref{fig:qualitative_results}. From the experiments, object-centric approach performs better due to its robustness. When a subject stands at the corner (Fig.~\ref{fig:corner}), the model still can classify the event of handover accurately due to its object-centric nature. In addition, when neither the object is detected nor it is in hand (Fig.~\ref{fig:unique_cases}), the model will not classify it as a handover event. Nonetheless, the model is not free from errors. Some failure cases are shown in Fig.~\ref{fig:failure_cases}. This may be due to the limited size of our custom dataset and some scenarios are not taken in account. The size of our custom dataset can be increased in the future involving more subjects to make it more robust for training. 

\iffalse
\setlength{\tabcolsep}{0.5mm}
\begin{table}[ht!]
\centering
\begin{tabular}{|c|c|c|c|}
\cline{3-4}
\multicolumn{2}{c}{} & \multicolumn{2}{|c|}{Correct Label} \\
\cline{3-4}
\multicolumn{2}{r|}{} & True & False \\
\hline
\multirow{2}{*}{\rotatebox[origin=c]{90}{Recognition Output}} & 
\raisebox{1cm}{\rotatebox{90}{True}} & \raisebox{-0.7mm}{\includegraphics[trim={1cm 1cm 18cm 1cm},clip,width=0.44\linewidth]{images/true_positive_matrix.png}} & \raisebox{-0.7mm}{\includegraphics[trim={1cm 1cm 18cm 1cm},clip,width=0.44\linewidth]{images/false_positive_matrix.png}} \\ 
\cline{2-4} & \raisebox{1cm}{\rotatebox{90}{False}} & \raisebox{-0.7mm}{\includegraphics[trim={1cm 1cm 18cm 1cm},clip,width=0.44\linewidth]{images/false_negative_matrix.png}} & 
\raisebox{-0.7mm}{\includegraphics[trim={1cm 1cm 18cm 1cm},clip,width=0.44\linewidth]{images/true_negative_matrix.png}} \\ \hline
\end{tabular}
\caption{Handover Gesture classification examples}
\label{tab:gt}
\end{table}
\setlength{\tabcolsep}{0.15cm}
\fi

\section{Conclusion}
We successfully demonstrated an approach for recognition of human-to-robot handovers with neural networks. The system takes in a single RGB frame and outputs a binary classification for the recognition of human-to-robot handovers. Several modules are incorporated, i.e. object detection, keypoints detection, head pose estimation and multi-layer perceptron. The system also performs computation on the body keypoints and this allows  handover situations to be detected regardless of the object's location and its surrounding environment. As a result, it becomes an object-centric frame. In general, the performance of the system is good to a large extent as the feature vectors can be classified accurately. Nonetheless, some errors can be observed during deployment due to a limited data set to train the multi-layer perceptron, and hence further development will be required. Since the pose of the human initiating handovers is only considered during the design of this system, no temporal information is considered by the system. Further, still images are sufficient to classify handover scenarios. Future work includes investigating the use of other communication cues (such as verbal and gaze), temporal information and detection in the presence of multiple objects. Ablation studies should be carried out to demonstrate the effect of each module in the system.

\bibliographystyle{IEEEtran}

\bibliography{refs}

\end{document}